\documentclass[conference]{IEEEtran}
\IEEEoverridecommandlockouts
\usepackage{cite}
\usepackage[hidelinks]{hyperref}
\usepackage{amsmath,amssymb,amsfonts}
\usepackage{graphicx}
\usepackage{textcomp}
\usepackage{xcolor}
\usepackage{mytodonotes}
\usepackage[inline]{enumitem}
\usepackage{myacronyms}
\usepackage{algorithm}
\usepackage{url}
\usepackage{algpseudocode}
\usepackage{amsthm}
\usepackage{cleveref}

\theoremstyle{plain}

\crefname{assumption}{assumption}{assumptions}
\Crefname{assumption}{Assumption}{Assumptions}
\theoremstyle{definition}
\newtheorem{definition}{Definition}
\crefname{definition}{definition}{definitions}
\Crefname{definition}{Definition}{Definitions}

\def\BibTeX{{\rm B\kern-.05em{\sc i\kern-.025em b}\kern-.08em
    T\kern-.1667em\lower.7ex\hbox{E}\kern-.125emX}}

\begin{document}

%% Tentative title
\title{
Discovering Collaboration from Novelty: Random Network Distillation for Clustered Federated Learning 
}

%Learning Who to Collaborate With: Random Network Distillation for Clustered Federated Learning

% Alternative titles:

% From Novelty to Collaboration: Autonomous Cluster Formation in Federated Learning
% Learning Autonomous Collaboration Structures with Random Network Distillation in Federated Learning
% Autonomous Federation Discovery via Random Network Distillation for Clustered Federated Learning
%Emergent Federations via Random Network Distillation in Federated Learning

%Emergent Federations via Random Network Distillation in Clustered Federated Learning

\author{\IEEEauthorblockN{Davide Domini}
\IEEEauthorblockA{%\textit{dept. name of organization (of Aff.)} \\
\textit{University of Bologna}\\
Cesena, Italy  \\
davide.domini@unibo.it \\
0009-0006-8337-8990
}
\and
\IEEEauthorblockN{Gianluca Aguzzi}
\IEEEauthorblockA{%\textit{dept. name of organization (of Aff.)} \\
\textit{University of Bologna}\\
Cesena, Italy  \\
gianluca.aguzzi@unibo.it \\
0000-0002-1553-4561}
\and
\IEEEauthorblockN{Ivana Dusparic}
\IEEEauthorblockA{%\textit{dept. name of organization (of Aff.)} \\
\textit{Trinity College Dublin}\\
Dublin, Ireland  \\
ivana.dusparic@tcd.ie \\
0000-0003-0621-5400}
\and
\IEEEauthorblockN{Danilo Pianini}
\IEEEauthorblockA{%\textit{dept. name of organization (of Aff.)} \\
\textit{University of Bologna}\\
Cesena, Italy  \\
danilo.pianini@unibo.it \\
0000-0002-8392-5409}
\and
\IEEEauthorblockN{Mirko Viroli}
\IEEEauthorblockA{%\textit{dept. name of organization (of Aff.)} \\
\textit{University of Bologna}\\
Cesena, Italy  \\
mirko.viroli@unibo.it \\
0000-0003-2702-5702}
}

\maketitle

\begin{abstract}
\acl{FL} often suffers under \acl{non-IID} data,
 where a single global model may fail 
 to represent the diversity of client distributions. 
\acl{CFL} mitigates this issue by training specialized models
 for groups of similar clients, 
 but existing approaches often couple cluster assignment with the main training loop,
 increasing computational and communication costs.
We propose a lightweight clustering approach based on \acl{RND}.
Each client trains a compact \acl{RND} predictor on its local data
 and uses its prediction error as a novelty signal to estimate similarity with other clients.
This enables the discovery of meaningful client groups before federated training, 
 without sharing raw data or repeatedly evaluating the main model. 
Crucially, 
 the resulting federations emerge from local novelty estimates at runtime, 
 making the method suitable for autonomous large-scale distributed systems
 where neither the number of clusters nor 
 the collaboration structure can be specified a priori.
Overall, 
 by decoupling clustering from learning, 
 the method provides a task-agnostic and efficient mechanism 
 for autonomous collaboration under \acl{non-IID} data.

\end{abstract}


% ================================================================
\section{Introduction}\label{sec:introduction}
% ================================================================
\ac{FL}~\cite{DBLP:conf/aistats/McMahanMRHA17} has emerged as a prominent paradigm
 for collaborative model training in privacy-sensitive distributed settings, 
 enabling devices to jointly learn shared models without exposing raw local data. 
While \ac{FL} achieves strong performance under homogeneous data conditions,
 real-world deployments frequently exhibit \ac{non-IID} data across devices,
 leading to degradation in model accuracy 
 and convergence~\cite{DBLP:conf/icde/LiDCH22,DBLP:journals/ftml/KairouzMABBBBCC21}.

This challenge becomes particularly severe in large-scale distributed systems 
 comprising numerous devices that must learn and adapt over time.
In many settings, 
 data heterogeneity is not random but structured. 
Spatially distributed systems provide a clear example:
 devices in spatial proximity tend to observe similar phenomena, 
 while devices in different regions exhibit statistically divergent 
 distributions~\cite{esterle2022deep}. 
This naturally gives rise to group-level \ac{non-IID} patterns, 
 where clusters of devices share similar local distributions internally, 
 while distributions differ significantly across clusters.
Furthermore, 
 in such autonomous settings, 
 manually specifying \emph{who should collaborate with whom} 
 or how many collaboration groups should exist is often unrealistic, 
 since the number and structure of latent clusters 
 are typically difficult to know in advance.

\acf{CFL}~\cite{liu2025survey} addresses this by partitioning devices into groups 
 and training a specialized model per cluster, 
 rather than a single global model. 
Algorithms such as IFCA~\cite{DBLP:journals/tit/GhoshCYR22} 
 and PSFL~\cite{DBLP:journals/iot/DominiFAVE26}
 have demonstrated that cluster-aware training 
 substantially improves accuracy under \ac{non-IID} conditions. 
However, 
 these methods share a fundamental limitation: 
 at each round, cluster assignment requires evaluating one 
 or more instances of the full model being trained. 
In IFCA, 
 every device must evaluate all cluster models on its local data 
 and select the one yielding the lowest loss; 
 in PSFL, devices continuously exchange and evaluate neighbors' models to maintain coherent federations. 
This per-round evaluation is computationally expensive, communication-intensive, 
 and may be prohibitive on resource-constrained devices.

In this paper, 
 we argue that clustering need not be entangled with the main learning loop. 
In many practical scenarios, 
 the cluster structure is either stable or changes slowly relative to the learning timescale. 
It is therefore sufficient to perform clustering once (or periodically) 
 using lightweight auxiliary models, entirely decoupled from the primary learning task.
Toward this goal, 
 we propose leveraging \acf{RND}~\cite{DBLP:conf/iclr/BurdaESK19}
 as a mechanism for estimating data uncertainty 
 and computing inter-device similarity for clustering. 
\ac{RND} was originally introduced to quantify epistemic uncertainty 
 in \ac{MARL} for exploration and \ac{TL}~\cite{DBLP:conf/ecai/CastagnaD23}, 
 where the prediction error of a randomly initialized fixed network serves as a proxy 
 for novelty with respect to an agent's experience. 
We find this formulation remarkably well-suited to the federated clustering problem: 
 devices can train small \ac{RND} models on their local data 
 and exchange them with neighbors to estimate distributional divergence--without 
 sharing any raw data---enabling a lightweight, 
 privacy-preserving clustering phase.

Concretely, 
 our approach introduces a modular, 
 plug-and-play pre-clustering phase applicable to existing \ac{FL} algorithms. 
Devices first train compact \ac{RND} networks (significantly smaller than the primary task model), 
 exchange them within the network, 
 and use the resulting uncertainty estimates to identify cluster membership. 
The primary learning phase then proceeds within the discovered clusters 
 using any standard aggregation algorithm such as FedAvg~\cite{DBLP:journals/corr/McMahanMRA16}. 
When data distributions shift over time, 
 the clustering phase can be re-triggered periodically.

To the best of our knowledge, 
 this is the first work to apply \ac{RND} for cluster discovery in \ac{FL}. 
We validate our approach on well-known computer vision benchmarks 
 augmented with synthetic noise to simulate feature skewness, 
 a realistic and underexplored form of \ac{non-IID} heterogeneity, 
 demonstrating that \ac{RND} uncertainty estimates effectively separate devices 
 into meaningful clusters prior to any model training.

% Commented out to save space
% The remainder of this paper is organized as follows: 
% \Cref{sec:background} provides background on Federated Learning, 
%  data heterogeneity, and Random Network Distillation. 
% % 
% \Cref{sec:motivation} presents the motivation 
%  and a reference scenario. 
% %
% \Cref{sec:formalization} formalizes the proposed method. 
% %
% \Cref{sec:evaluation} reports the experimental evaluation. 
% %
% Finally, 
%  \Cref{sec:conclusions} concludes the paper 
%  and outlines directions for future work.

% ================================================================
\section{Background and Related Works}\label{sec:background}
% ================================================================

\subsection{Federated Learning}

% \begin{figure}
%     \centering
%     \includegraphics[width=0.75\columnwidth]{figures/fl.pdf}
%     \caption{A visual representation of client-server federated learning. }
%     \label{fig:fl}
% \end{figure}

\acf{FL}~\cite{DBLP:conf/aistats/McMahanMRHA17} is a distributed learning paradigm 
 in which multiple clients collaboratively train machine learning models 
 while keeping their raw data locally stored. 
In the standard server-based formulation, % (\Cref{fig:fl}), 
 clients perform local optimization on private datasets
 and periodically send model updates to a central server, 
 which aggregates them to produce an updated global model. 
The most common aggregation strategy is FedAvg~\cite{DBLP:journals/corr/McMahanMRA16}, 
 where the global model is obtained by averaging the parameters or updates received 
 from the participating clients.

While standard \ac{FL} is effective when client data are sufficiently homogeneous,
 its performance can degrade under \acf{non-IID} data. 
In such settings, 
 local updates may be biased toward different client-specific objectives, 
 making their aggregation less stable and reducing the quality 
 of the resulting global model~\cite{liu2025survey,DBLP:journals/ftml/KairouzMABBBBCC21}.
This issue has motivated several extensions of \ac{FL} aimed at improving 
 robustness under statistical heterogeneity.

\acf{CFL} addresses data heterogeneity by replacing the single global model 
 with multiple specialized models. 
Clients are partitioned into clusters according to some notion of similarity, 
 and each cluster trains a separate federated model. 
This formulation is particularly useful when the client population is composed 
 of groups with internally similar but mutually different data distributions. 
Representative \ac{CFL} approaches include IFCA~\cite{DBLP:journals/tit/GhoshCYR22},
 which alternates between assigning clients to the model that best fits 
 their local data and updating cluster-specific models, 
 and proximity-aware approaches such as PSFL~\cite{DBLP:journals/iot/DominiFAVE26}, 
 where client federations are formed and maintained according to local 
 similarity estimates among neighboring devices.

\subsection{Data Heterogeneity}

\begin{figure}
    \centering
    \includegraphics[width=0.5\columnwidth]{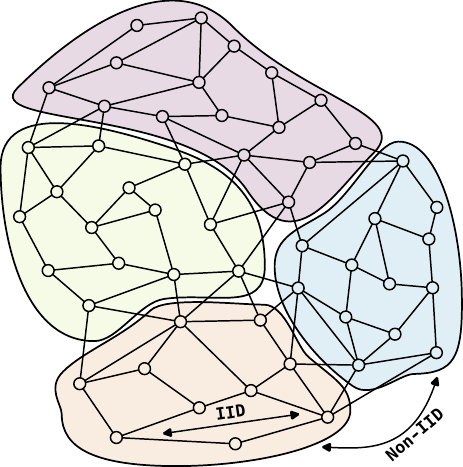}
    \caption{Clustered data heterogeneity. 
    Each node represents a device, and each background region denotes a different underlying feature distribution. Devices within the same cluster collect data from the same distribution, 
    whereas devices across different clusters observe non-IID data.}
    \label{fig:subregions}
\end{figure}

Data heterogeneity is one of the main challenges in \ac{FL}~\cite{DBLP:conf/middleware/NilssonSUGJ18}, 
 since clients usually collect data through different devices, sensors, users, 
 or environmental conditions rather than sampling from a single shared distribution. 
In real-world deployments,
 two clients may observe the same task from different perspectives: 
 for instance, images of the same classes may be captured under different lighting conditions,
 cameras, backgrounds, styles, or noise patterns. 
Similarly, 
 sensing devices deployed in different environments may measure the same type 
 of phenomenon while being exposed to different local dynamics.

Such heterogeneity can manifest in multiple forms, 
 depending on how data are partitioned among 
 clients~\cite{DBLP:conf/icde/LiDCH22,DBLP:journals/ftml/KairouzMABBBBCC21}.
Commonly studied types of data skew include:
\begin{enumerate*}[label=(\roman*)]
  \item \emph{feature skew}, where all clients share the same label space but differ in their feature distributions (\textit{e.g.}, handwritten character recognition with varying writing styles);
  \item \emph{label skew}, in which each client observes only a subset of the global classes; and
  \item \emph{quantity skew}, where clients possess highly imbalanced amounts of local data.
\end{enumerate*}
Feature skew is therefore especially relevant when the semantic task is shared across clients, 
 but the way data are observed varies across devices or contexts.

As discussed in~\Cref{sec:introduction}, 
 data skewness in \ac{FL} can be structured rather than arbitrary. 
In this work,
 we formalize this setting as clustered feature skew, 
 where clients belong to latent clusters 
 with aligned feature distributions within each cluster 
 and distinct distributions across clusters, as illustrated in~\Cref{fig:subregions}.

% In \ac{CFL}, 
%  data heterogeneity can exhibit a structured form: 
%  clients can be grouped into latent clusters, 
%  where clients within the same cluster observe data generated under similar conditions, 
%  while clients in different clusters observe different feature distributions---as 
%  shown in~\Cref{fig:subregions}. 
% % 
% Accordingly, 
%  the heterogeneity considered in this paper arises from feature skew 
%  with an explicit clustered structure.
%  This setting is formalized as follows.

\begin{definition}[Clustered Feature Skew]\label{def:clustered-feature-skew}
Let $\mathcal{D}=\{d_1,\ldots,d_N\}$ be a set of clients partitioned into
 $k$ latent clusters $C_1,\ldots,C_k$. 
Each client $d_i$ is associated with a local data distribution $P_i(x,y)$,
 where $x$ is the input and $y$ is the corresponding label.
We say that the clients satisfy \emph{clustered feature skew} 
 if all clients share the same label marginal distribution $P(y)$,
 while their class-conditional feature distributions are 
 identical within each cluster and different across clusters.
In other words,
 all clients have the same label proportions,
 but clients in different clusters represent the same labels using different feature distributions.
Formally, 
 for any two clients $d_i,d_j \in C_q$, holds: 
\[
P_i(y)=P_j(y),
\qquad
P_i(x\mid y)=P_j(x\mid y),
\]
whereas for any two clients $d_i \in C_q$ and $d_j \in C_r$, with $q \neq r$:
\[
P_i(y)=P_j(y),
\qquad
P_i(x\mid y)\neq P_j(x\mid y).
\]
\end{definition}

\subsection{Random Network Distillation}

\acf{RND} is a lightweight method for estimating the novelty of an input 
 with respect to a set of previously observed examples~\cite{DBLP:conf/iclr/BurdaESK19}. 
This method is based on the observation that neural networks tend to produce 
 lower prediction errors on inputs similar to those seen during training, 
 and higher errors on unfamiliar or distributionally different inputs. 
Therefore, 
 the prediction error of a network trained on past data 
 can be used as a proxy for novelty, 
 or equivalently as an estimate of uncertainty on previously unseen samples.

Basically, 
 \ac{RND} consists of two neural networks. 
A target network, 
 randomly initialized and kept fixed, defines a mapping (or embedding):
\begin{equation}
    f: \mathcal{X} \rightarrow \mathbb{R}^{k},
\end{equation}
while a predictor network
\begin{equation}
    \hat{f}(\cdot;\theta): \mathcal{X} \rightarrow \mathbb{R}^{k}
\end{equation}
is trained to approximate the target network \emph{on the available data}.
Given a data distribution $\mathcal{B}$,
 the predictor is optimized by minimizing the expected mean squared error
\begin{equation}\label{eq:rnd}
    \theta^{*}
    =
    \arg\min_{\theta}
    \mathbb{E}_{x \sim \mathcal{B}}
    \left[
        \left\|
            \hat{f}(x;\theta) - f(x)
        \right\|_2^2
    \right].
\end{equation}
Once trained,
 the \ac{RND} uncertainty score of a sample $x$ is computed as
\begin{equation}
    u(x)
    =
    \left\|
        \hat{f}(x;\theta^{*}) - f(x)
    \right\|_2^2 .
\end{equation}
Intuitively,
the two network will produce similar outputs for inputs that are similar to the training data,
while they will diverge on novel inputs.
Thus, low values of $u(x)$ indicate that $x$ is similar to the data used to train the predictor,
 whereas high values suggest that $x$ is novel or comes from a different distribution.

In this work, 
 we exploit this property to estimate similarity between clients in \ac{FL}. 
A predictor trained on the local data of one client is expected 
 to yield low uncertainty on data drawn from clients with similar feature distributions,
 and higher uncertainty on data from clients belonging to different clusters.
\ac{RND} therefore provides a compact and task-agnostic mechanism 
 to compare local data distributions without exchanging raw data 
 or evaluating the main federated model.

% ================================================================ 
\section{Motivation and Reference Scenario}\label{sec:motivation}
% ================================================================
 
Consider a set of clients collaboratively training an image recognition model
 from photos collected on mobile devices.
Clients may share the same semantic task--recognizing cars, buildings,
 people, or street signs--while observing the world under substantially
 different visual conditions.
Photos taken in different cities naturally differ in illumination, weather,
 architectural style, or sensor characteristics.
As a result, 
clients from the same city tend to benefit from collaborating with one another, 
 whereas aggregating their updates with clients from visually
 different environments may produce a model that serves no group well.

A symmetric setting arises in distributed sensing.
Environmental monitoring devices deployed across heterogeneous areas may all
 measure the same physical phenomenon, 
 yet under markedly different local conditions.
Sensors placed near a coastal area, an urban district, or an industrial zone
 observe different noise patterns, dynamics, and baseline statistics.
These scenarios expose two intertwined challenges for autonomous distributed systems, namely:
\begin{enumerate*}[label=(\roman*)]
    \item determining who should collaborate with whom; and
    \item allowing the corresponding federations to emerge without requiring 
    their number to be specified externally.
%    since the structure of latent clusters is difficult to know in advance.
\end{enumerate*}

\ac{CFL} addresses the first challenge by training specialized models 
 for groups of compatible clients.
However, 
 existing approaches only partially address the autonomous discovery of such groups.
In IFCA, 
 the number of clusters is a hyperparameter that must be fixed before training; 
 an incorrect choice may lead to suboptimal assignments 
 and affect the quality of the final models---as shown in~\cite{DBLP:journals/iot/DominiFAVE26}.
Moreover, 
 cluster assignment is coupled with the primary learning task, 
 since clients must evaluate full-size cluster models on their local data.
PSFL relaxes the need to predefine a global number of clusters 
 by allowing federations to emerge from local interactions, 
 but still estimates client compatibility through the full task model.
Thus, in both cases, 
 discovering collaboration structures remains tied to
 repeated evaluations of the model being trained.

In the scenarios above, 
 this repeated coupling is largely unnecessary.
The relevant client groups reflect relatively stable data-generation conditions,
 such as the physical environment in which sensors are deployed,
 that are unlikely to shift dramatically between rounds.
Clustering can therefore be treated as a \emph{separate} 
 and cheaper preliminary step, performed once before federated training begins, 
 or re-triggered occasionally when distributions change.

The key challenge is then to estimate whether two clients observe compatible data distributions, 
 without exchanging raw data and without relying on the primary model.
Rather than asking whether a full task model trained 
 by one client generalizes to another client's data, we reframe the problem:
 does one client's data appear \emph{novel} with respect 
 to another client's local experience?

% ================================================================
\section{Method Formalization}\label{sec:formalization}
% ================================================================

\subsection{Problem Formulation}

Let $\mathcal{D}=\{d_i\}_{i=1}^{N}$ be the set of $N$ devices participating in the \ac{FL} process.
Each device $d_i$ holds a local dataset $B_i$:
\begin{equation}
 B_i=\{z_{ij}\}_{j=1}^{m_i},
 \qquad
 z_{ij}=(x_{ij},y_{ij})\sim P_i .
\end{equation}
where $x_{ij}$ is the $j$-th input feature vector on device $d_i$, $y_{ij}$ is the corresponding label,
and $m_i$ is the number of local samples.
Here, each local distribution $P_i$ satisfies clustered
feature skew, as defined in \Cref{def:clustered-feature-skew}.

The devices are partitioned into groups (clusters) $C_1, \dots, C_k$,
with $C_q \cap C_r = \emptyset$ for $q \neq r$ (no overlap)
and $\bigcup_{q=1}^{k} C_q = \mathcal{D}$ (all devices belong to some cluster),
however,
 this partition is \emph{not known} to the devices at learning time. 
Formally,
 the goal is to learn a mapping $\mathcal{M}$ from the set of devices 
 to a partition of $\mathcal{D}$:
\begin{equation}
    \mathcal{M}: \mathcal{D} \rightarrow \mathcal{F} \subseteq \mathcal{P}(\mathcal{D})
\end{equation}
where $\mathcal{P}(\mathcal{D})$ denotes the power set of $\mathcal{D}$, 
 and $\mathcal{M}$ produces a set of federations $\mathcal{F} = \{F_1, \dots, F_f\}$, 
 where each federation $F_q \subseteq \mathcal{D}$ represents a group of devices 
 that will collaboratively train a specialized model. 
The goal is that each discovered federation $F_q$ matches as closely as possible 
 the corresponding true cluster $C_q$. 
Formally, 
 we aim to find $\mathcal{M}$ that minimizes the \emph{misclassification rate}:
\begin{equation}
    \min_{\mathcal{M}} \frac{1}{N} \min_{\pi \in \Pi_k} 
    \sum_{q=1}^{f} \left| F_q \triangle C_{\pi(q)} \right|
\end{equation}
where $\Pi_k$ is the set of all permutations over $\{1, \dots, k\}$, 
 $\triangle$ denotes the symmetric difference between two sets, 
 and $|\cdot|$ is the cardinality. 
The permutation $\pi$ accounts for the fact 
 that federation labels are arbitrary. 
Once $\mathcal{M}$ is determined, 
 each federation $F_q \in \mathcal{F}$ independently runs a federated learning process, 
 training a specialized model $\omega_q$ optimized 
 for the shared data distribution within that group.

\subsection{Novelty Driven Clustering}

This section details the proposed novelty driven federation discovery method, 
 whose device-side and server-side procedures
%\footnote{
% A completely distributed version of the method is also possible
% based on leader election or gossip protocols,
% but is not explored in this work.
%}
are summarized in
 Algorithms~\ref{alg:device-rnd} and~\ref{alg:server-rnd}, respectively.

Each device $d_i$ trains a local RND predictor $\hat{f}_i(\cdot;\theta_i)$ 
 by minimizing the expected prediction error with respect to a shared, 
 randomly initialized target network $f$, as formalized in~\Cref{eq:rnd}.
Once trained, each device broadcasts its predictor $\hat{f}_i(\cdot;\theta_i^*)$ 
 to all other devices.
Each device $d_i$ then evaluates every received predictor $\hat{f}_j$ 
 on its own local data $B_i$, 
 obtaining a novelty score that quantifies the distributional divergence 
 between $d_j$ and $d_i$:
\begin{equation}
    s_{ij} = \mathbb{E}_{x \sim B_i} 
    \left[ \left\| \hat{f}_j(x;\theta_j^*) - f(x) \right\|_2^2 \right],
    \quad \forall j \in \{1, \ldots, N\}.
\end{equation}
Low values of $s_{ij}$ indicate that $d_j$'s predictor 
 generalizes well to $d_i$'s data, 
 suggesting similar underlying distributions; 
 high values indicate distributional divergence. 
Each device $d_i$ then reports its row of scores $\{s_{ij}\}_{j=1}^{N}$ 
 to the server, which reconstructs the full similarity matrix 
 $S \in \mathbb{R}^{N \times N}$ where $S[i,j] = s_{ij}$,
 without ever accessing raw local data.

\paragraph{Adaptive Federation Discovery}
%rather than fixing the number of federations a priori,
 the partition $\mathcal{F}$ can be derived directly 
 from the observed novelty scores via a compatibility threshold.
Each device $d_i$ uses self-novelty $s_{ii}$ as reference baseline,
 reflecting the residual prediction error on its own data after local training.
Device $d_j$ is compatible with $d_i$ if
 its cross-novelty score does not exceed the self-novelty 
 by more than a relative margin:
\begin{equation}
    d_j \sim d_i \iff s_{ij} \leq s_{ii} + \epsilon \cdot \sigma_i,
\end{equation}
where $\sigma_i$ is the standard deviation of $\{ s_{ij} \}_{j=1}^{N}$ 
 and $\epsilon \geq 0$ is a tolerance hyperparameter.
The number of federations emerges naturally 
 as the number of distinct compatibility groups identified across devices,
 rather than being specified in advance,
 and is expected to converge to the true number of clusters $k$.
Once the partition $\mathcal{F} = \{F_1, \ldots, F_f\}$ is determined, 
 each federation $F_q$ independently runs a federated learning process.
% using a standard aggregation algorithm such as FedAvg.

\paragraph{Clustering Frequency}
in stationary environments, 
 the clustering phase needs to be performed only once 
 prior to any federated training, since the data-generation conditions 
 that define the cluster structure are not expected to change.
When data distributions evolve over time~\cite{DBLP:journals/corr/abs-2606-18003},
% (e.g., due to sensor drift, seasonal variation, or changes in user behavior),
 the clustering phase can be re-triggered periodically. 
We introduce a re-clustering interval $\tau$ as a hyperparameter, 
 expressed in number of federated rounds, 
 such that the full \ac{RND}-based clustering procedure is repeated every $\tau$ rounds.
%
% Setting $\tau = +\infty$ recovers the stationary case, 
%  while smaller values of $\tau$ allow the system to adapt to distributional shifts 
%  at the cost of additional communication and computation.

\paragraph{Decentralized Extension}
while the formulation above assumes a central server that collects all 
 predictors and computes the full similarity matrix $S$,
 the approach extends naturally to decentralized settings.
A fully decentralized implementation can be realized by selecting the ``server'' via leader election,
or by propagating the predictors across the network through a gossip protocol.
Alternatively,
each device $d_i$ may exchange predictors only with
 its neighbors $\mathcal{N}(i)$, obtaining a partial view of the similarity matrix 
 restricted to reachable peers.
Cluster assignments can then be derived locally and reconciled into a 
 globally consistent partition via a consensus protocol among neighboring devices,
 similarly to how decentralized aggregation has been approached in prior 
 work~\cite{DBLP:journals/lmcs/DominiAEV26,DBLP:journals/ssrn/sparseful}.
This decentralized variant preserves the core properties of the approach
 (i.e., no raw data exchange, decoupling from the main learning loop)
 while removing the assumption of a central coordinator.

\begin{algorithm}[t]
\caption{Device-Side RND Novelty Estimation}
\label{alg:device-rnd}
\begin{algorithmic}[1]
\Require Device $d_i$ with local dataset $\mathcal{B}_i$
\Require Shared random target network $f$
\Ensure Local predictor $\hat{f}_i$ and novelty row $\mathbf{s}_i$

\State $\hat{f}_i \gets \Call{TrainRNDPredictor}{\mathcal{B}_i, f}$

\State $\Call{SendPredictor}{d_i, \hat{f}_i}$

\State $\{\hat{f}_j\}_{j=1}^{N} \gets \Call{ReceivePredictors}{d_i}$

\ForAll{$\hat{f}_j \in \{\hat{f}_1,\dots,\hat{f}_N\}$}
    \State $s_{ij} \gets \Call{ComputeNovelty}{\mathcal{B}_i, \hat{f}_j, f}$
\EndFor

\State $\mathbf{s}_i \gets (s_{i1}, \dots, s_{iN})$

\State $\Call{SendNoveltyRow}{d_i, \mathbf{s}_i}$

\end{algorithmic}
\end{algorithm}

\begin{algorithm}[t]
\caption{Server-Side RND-Based Federation Discovery}
\label{alg:server-rnd}
\begin{algorithmic}[1]
\Require Devices $\mathcal{D} = \{d_1,\dots,d_N\}$
\Require Compatibility tolerance $\epsilon \geq 0$
\Require Re-clustering interval $\tau$
\Ensure Discovered federations $\mathcal{F}$

\State $r \gets 0$

\While{training is not completed}

    \If{$r = 0$ \textbf{or} $r \bmod \tau = 0$}

        \State $\{\hat{f}_i\}_{i=1}^{N} \gets \Call{CollectPredictors}{\mathcal{D}}$

        \State $\Call{DistributePredictors}{\mathcal{D}, \{\hat{f}_i\}_{i=1}^{N}}$

        \State $\{\mathbf{s}_i\}_{i=1}^{N} \gets \Call{CollectNoveltyRows}{\mathcal{D}}$

        \State $S \gets \Call{BuildNoveltyMatrix}{\{\mathbf{s}_i\}_{i=1}^{N}}$

        \State $\mathcal{F} \gets \Call{ExtractFederations}{S}$

    \EndIf

    \State $\Call{RunFederatedLearning}{\mathcal{F}}$

    \State $r \gets r + 1$

\EndWhile

\State \Return $\mathcal{F}$

\end{algorithmic}
\end{algorithm}

% ================================================================
\section{Experimental Evaluation}\label{sec:evaluation}
% ================================================================

\subsection{Experimental Setup}
The experimental evaluation is designed to validate whether
 the proposed novelty driven procedure can recover the latent groups 
 induced by clustered feature skew. 

We used CIFAR-10~\cite{krizhevsky2009learning},
a well-known benchmark dataset for computer vision
tasks, and synthetically partitioned it to reproduce the case study
described in Section~\ref{sec:formalization}. 
The partitioning was performed using ProFed~\cite{DBLP:journals/jors/DominiIIAEV26}. 
Given a number of groups $k$, 
 the dataset was divided into $k$ subsets 
 while preserving the same label distribution in each group. 
Then, 
 a different Gaussian noise distribution was applied to the images of each group,
 thereby inducing group-specific feature skew 
 while keeping the underlying classification task unchanged.
In our experiments, 
 we set $k=4$ and assigned three devices to each group, 
 resulting in a total of $N=12$ devices. 

For the RND predictors, 
 we used a lightweight CNN composed of two convolutional layers 
 followed by a flattening layer. 
After local training, 
 each device evaluates both its own predictor 
 and the predictors trained by all other devices on its local data,
 producing the novelty scores $s_{ij}$ used to construct the novelty matrix. 
We then measured: 
\begin{enumerate*}[label=(\roman*)]
    \item the resulting novelty scores, 
     to assess whether devices within the same group exhibit similar residuals; and
    \item the total machine-time\footnote{
     taken on an Intel Core i7 8700K, 64GB RAM, nVidia RTX 4070.
    } spent to perform clustering,
     including local RND training and inference with all received predictors.
\end{enumerate*}

To contextualize the overhead of the proposed method, 
 we also measured the clustering cost of IFCA. 
In this case, 
 cluster assignment relies on evaluating the full task model 
 rather than a lightweight auxiliary predictor. 
We used a ResNet-18~\cite{DBLP:conf/cvpr/HeZRS16} as the full classification model, 
 since this architecture is commonly adopted for CIFAR-10 
 and is known to provide strong performance on this dataset. 

Each experiment was repeated with $10$ different random seeds 
 to account for stochasticity and avoid cherry-picking favourable runs. 
The reported results correspond to averages over these independent repetitions. 
The experimental code is publicly 
 available\footnote{\url{https://github.com/domm99/experiments-2026-uncertainty-based-clustered-fl}}.

\subsection{Discussion}

\begin{figure}[t]
    \centering
    \includegraphics[width=0.85\columnwidth]{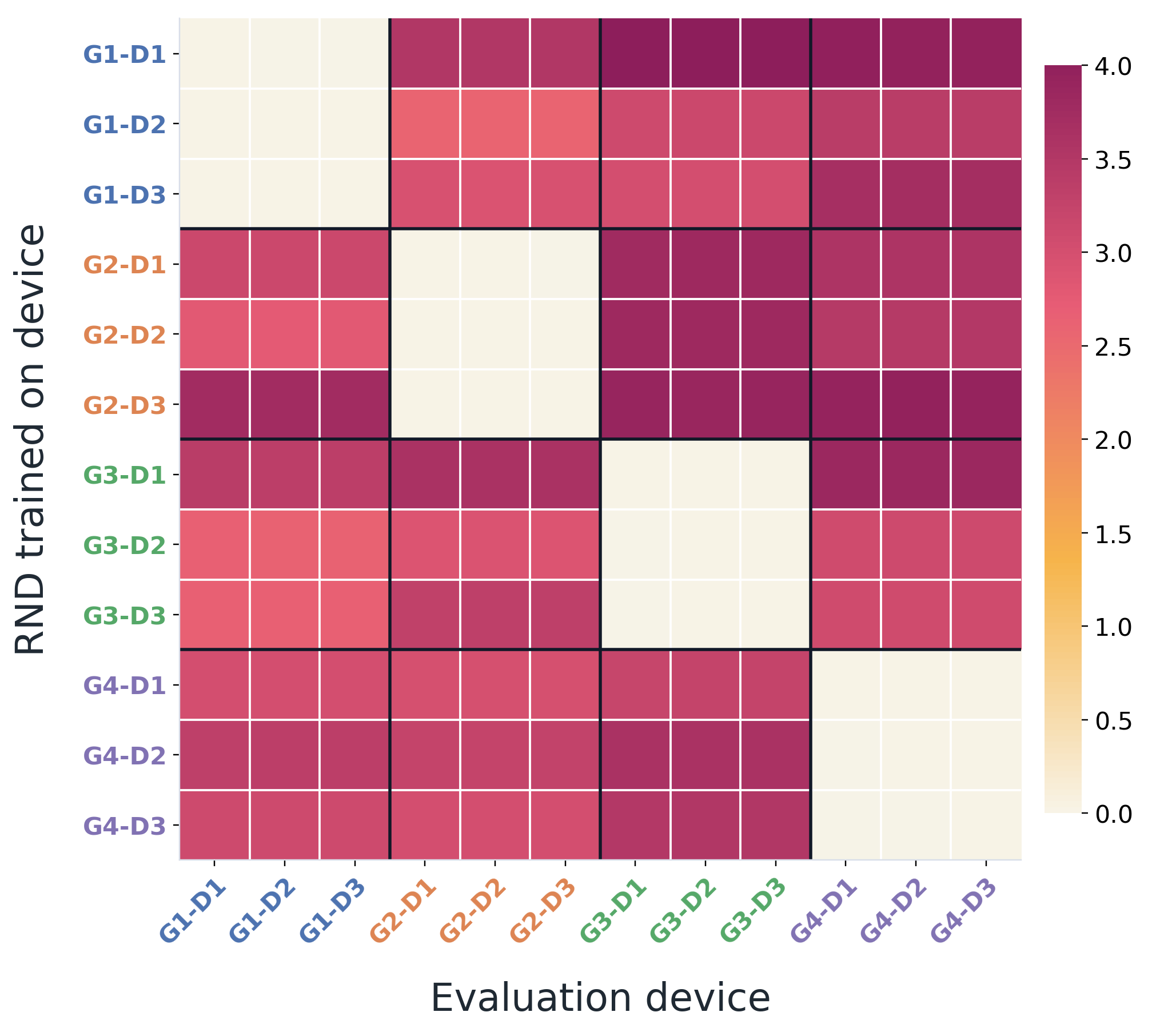}
    \caption{
     Difference between self-novelty $s_{ii}$ and cross-novelty $s_{ij}$ across devices. 
     The colors of the device labels indicate the true clusters used to generate the experimental data.
     Near-zero values within diagonal blocks indicate that devices 
     in the same group induce similar RND residuals.
    }
    \label{fig:chart-difference}
\end{figure}

\begin{figure}[t]
    \centering
    \includegraphics[width=0.8\columnwidth]{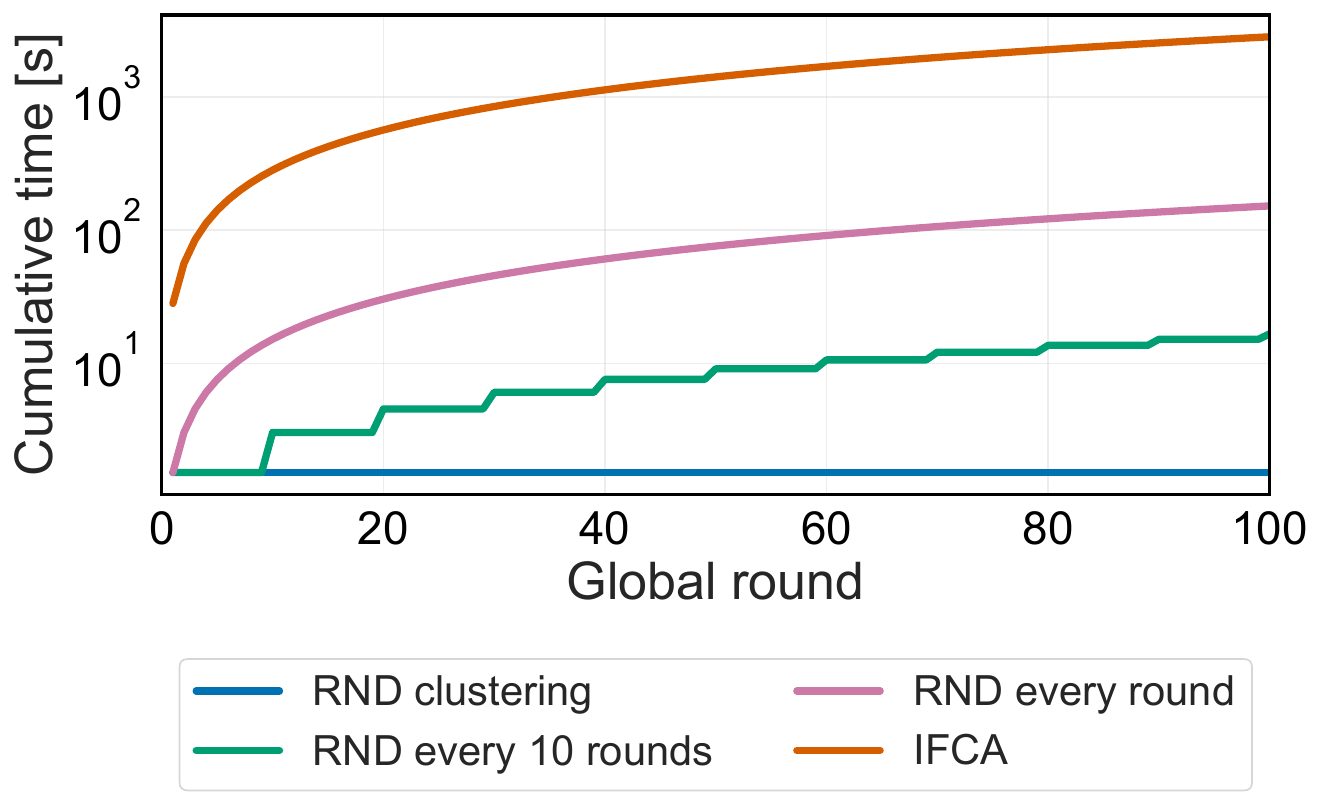}
    \caption{
     Clustering overhead over federated rounds. 
     Compared with IFCA, RND-based clustering reduces cost by re-clustering less frequently 
     and by relying on small auxiliary predictors instead of full task models.
    }
    \label{fig:chart-time}
\end{figure}

Figure~\ref{fig:chart-difference} reports the results of the first experiment, 
 where each device evaluates the \ac{RND} predictors trained 
 by all other devices on its own local data. 
In particular, 
 the figure shows the difference between the local residual $s_{ii}$ 
 and the cross-novelty score $s_{ij}$. 
The diagonal block structure clearly highlights that, 
 for devices belonging to the same group, 
 this difference tends to be close to zero. 
This indicates that predictors trained within 
 the same group produce similar residuals, 
 and therefore generalize to each other's data.
Conversely, 
 the difference remains higher across groups,
 showing that \ac{RND} novelty scores capture the distributional divergence 
 induced by feature skew. 
These results support the use of \ac{RND} as a lightweight proxy
 for identifying which devices should collaborate.

An important aspect of the proposed method is that 
 the number of federations does not need to be specified a priori. 
This is in contrast with approaches such as IFCA, 
 where the number of clusters is typically fixed before training. 
In highly distributed scenarios, 
 estimating this number can be difficult  and an incorrect choice 
 may affect the quality of the final models.
Instead, in our approach, 
 federations are derived from the structure of the novelty matrix 
 through the compatibility criterion,
 allowing groups of similar devices to emerge from the observed data.

Figure~\ref{fig:chart-time} reports the overhead associated 
 with the clustering phase. 
The proposed \ac{RND}-based approach exhibits substantially lower cost than IFCA, 
 since clustering relies on compact auxiliary predictors rather 
 than on the full task model used for \ac{FL}. 
Moreover, 
 \ac{RND}-based clustering does not need to be executed at every global round: 
 in stationary settings it can be performed once before training, 
 while in non-stationary settings it can be repeated periodically.
Even in the most conservative configuration,
 where \ac{RND} clustering is repeated at every round, 
 the overhead remains about one order of magnitude lower 
 than IFCA in our experiments.

A limitation of the centralized formulation 
 is that each device evaluates the predictors received from all other devices. 
As the number of devices grows, 
 this all-to-all evaluation may reduce the overhead gap with
 respect to full-model approaches. 
However, 
 this is not intrinsic to the method. 
In large-scale distributed deployments, 
 a decentralized implementation can restrict predictor exchange 
 and evaluation to local neighborhoods. 
In that case, 
 each device only evaluates the predictors of nearby peers, 
 whose number is typically much smaller than the total population size. 

% ================================================================
\section{Conclusions and Future Work}\label{sec:conclusions}
% ================================================================

In this paper, 
 we introduced an \ac{RND}-based clustering mechanism for \acl{CFL} 
 under structured feature skew.
The proposed approach decouples federation discovery from the main learning process 
 by relying on lightweight auxiliary predictors instead of full task models. 
As a result, 
 devices can infer suitable collaboration groups from novelty estimates, 
 without sharing raw data and without requiring 
 the number of clusters to be fixed a priori.

Preliminary experiments on synthetically partitioned CIFAR-10 show 
 that \ac{RND} residuals expose the latent clustered structure of the data 
 and can separate devices affected by different feature perturbations. 
The results also indicate a substantially lower 
 clustering overhead compared with IFCA, 
 especially when clustering is performed once or periodically 
 rather than at every global round.

Future work will extend the evaluation to larger populations of devices,
 additional forms of \ac{non-IID} data, and real-world distributed sensing scenarios. 
We also plan to investigate fully decentralized variants, 
 where devices exchange \ac{RND} predictors only with local neighbors.

\section*{Acknowledgments}
The authors used Generative AI tools for grammar 
 and language editing assistance throughout the manuscript. 
All AI-assisted text was reviewed and edited by the authors, 
 who take full responsibility for the content of this paper.

\bibliographystyle{IEEEtran}
\bibliography{IEEEexample}

\end{document}